\documentclass{bmvc2k}

\usepackage{multirow}
\usepackage{graphicx}
\usepackage{booktabs}
\usepackage[font=small]{caption}
\usepackage{hyperref}

\def\eg{\emph{e.g}.} 
\def\ie{\emph{i.e}.}

\def\etal{\emph{et al}.}

\title{Robust Action Segmentation from Timestamp Supervision}

\addauthor{Yaser Souri*}{yasersouri@microsoft.com}{1, 2}
\addauthor{Yazan Abu Farha*}{yabufarha@birzeit.edu}{1, 3}
\addauthor{Emad Bahrami*}{bahrami@iai.uni-bonn.de}{1}
\addauthor{Gianpiero Francesca}{gianpiero.francesca@toyota-europe.com}{4}
\addauthor{Juergen Gall}{gall@iai.uni-bonn.de}{1}

\addinstitution{
 Computer Vision Group\\
 University of Bonn\\
 Bonn, Germany
}
\addinstitution{
 Microsoft\\
 Redmond, United States
}
\addinstitution{
 Birzeit University\\
Birzeit, West Bank, Palestine
}
\addinstitution{
Toyota Motor Europe\\
 Brussels, Belgium\\
 \vspace{2mm}
 * indicates equal contribution
}

\runninghead{Souri \etal}{Robust Action Segmentation from Timestamp Supervision }

\begin{document}

\maketitle

\begin{abstract}
    Action segmentation is the task of predicting an action label for each frame of an untrimmed video. As obtaining annotations to train an approach for action segmentation in a fully supervised way is expensive, various approaches have been proposed to train action segmentation models using different forms of weak supervision, \eg, action transcripts, action sets, or more recently timestamps. Timestamp supervision is a promising type of weak supervision as obtaining one timestamp per action is less expensive than annotating all frames, but it provides more information than other forms of weak supervision. However, previous works assume that every action instance is annotated with a timestamp, which is a restrictive assumption since it assumes that annotators do not miss any action. 
In this work, we relax this restrictive assumption and take missing annotations for some action instances into account. We show that our approach is more robust to missing annotations compared to other approaches and various baselines.
\end{abstract}

\section{Introduction}
\label{sec:intro}

\begin{figure}[t]
   \centering
      \includegraphics[trim={0 0 0 5pt},clip,width=\columnwidth]{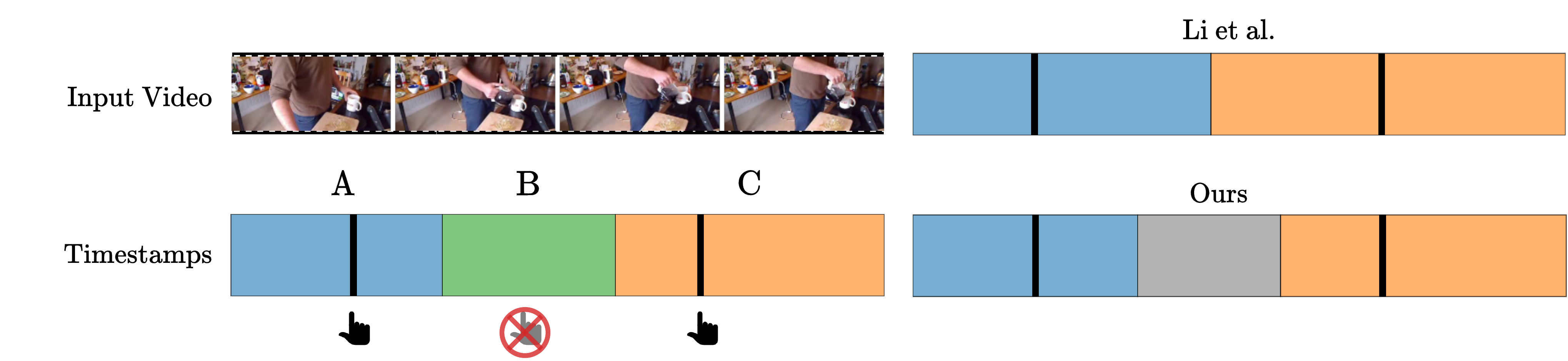}
   \caption{For an input video with 3 actions A, B, and C, the annotator has correctly annotated a timestamp (black line) for actions A and C, but action B has been missed. In this setting, Li \etal~\cite{li2021temporal} assume that all actions are annotated and generate a labeling from the annotated timestamps that is incorrect for the entire duration of action B. We propose an approach that takes into account the possibility of missing timestamps and correctly ignores the frames (gray) corresponding to the missed action B.}
   \label{fig:teaser}
\end{figure}

Action segmentation is an important task for many applications like home monitoring systems~\cite{toyota_smarthome_untrimmed},
worker monitoring and guidance in the assembly line~\cite{retrocausal}, or tutorial generation from instructional videos~\cite{cross_task, coin_dataset}. It requires to identify for each frame in an untrimmed video what action is happening. Recently, fully supervised approaches for action segmentation~\cite{farha2019ms,li2020ms,fifa2021,yi2021asformer} have achieved very good results. However, a major bottleneck that prevents the widespread adoption of action segmentation technologies is the cost of obtaining annotations for fully supervised learning. Annotating every instance of all actions of interest with their exact temporal boundaries is a labor-intensive endeavor. In particular, the start and end of an action is subjective and can lead to annotation inconsistencies.

To address these issues, several approaches have been proposed that train networks for action segmentation using only weak supervision, \eg, in form of action sequences without any time information~\cite{mucon2021,richard2018neuralnetwork,li2019weakly} or video tags~\cite{li2020set,fayyaz2020set,richard2018action}. 

While weakly supervised learning significantly reduces the annotation effort, these approaches still perform considerably worse than fully supervised approaches. Recently, \cite{moltisanti2019action} proposed to annotate only one frame, called timestamp, per action and this type of annotation has been also used to train a network for action segmentation in~\cite{li2021temporal}. Timestamp supervision is a form of weak supervision where the annotator annotates only a single frame and its class for each action in a video. While it requires to annotate only a very small fraction of the frames, this type of weak supervision provides a form of temporal information that is crucial for action segmentation and that is missing in other forms of weak supervision.
While the cost of obtaining timestamps is only marginally higher than other annotations for weakly supervised learning, \cite{li2021temporal} showed that timestamp supervision results in a much higher action segmentation performance.

Despite the promising results, \cite{li2021temporal} assumes that every action instance in a video is annotated by a timestamp since the approach detects an action change between two consecutive timestamps and labels the frames before and after the detected change based on the action classes of the corresponding timestamps as it is illustrated Figure~\ref{fig:teaser}. Annotators, however, can easily miss an action. In this case, the frames of the missed action are wrongly annotated. Since the generated labels are then used for training a network for action segmentation, the accuracy of the network drastically decreases when only 80-90\% of all action instances are annotated as shown in Figure~\ref{fig:gap}.      

In this paper, we thus address this limitation and propose an approach that is designed to deal with missing timestamps. In contrast to \cite{li2021temporal}, our approach can label frames as unknown as illustrated in Figure~\ref{fig:teaser}. These frames will then be ignored when the network for action segmentation is trained. To generate a labeling of a training video from a few annotated timestamps, we optimize the segment boundaries for all timestamps jointly. During the optimization, we expand the segment boundaries from the timestamps in each direction such that the network is confident that the frames within a segment belong to the same
class as the timestamp and such that the number of frames that are unknown is minimized. This is also illustrated in Figure~\ref{fig:parameters}.    

We evaluate our approach on two action segmentation benchmarks and show that our approach is considerably more robust to missing timestamps compared to \cite{li2021temporal} and various baselines. On the 50Salads~\cite{stein2013combining} dataset, the accuracy of our approach drops by only $4.5\%$ when the percentage of annotated timestamps is reduced from $95\%$ to $70\%$ compared to a drop of $18.2\%$ for \cite{li2021temporal}.
We further evaluate our approach on two datasets for action localization.   
While the goal of the proposed approach is to increase the robustness to missing timestamps, the approach also performs well when all action instances are annotated.

\section{Related Work}
\paragraph{Fully Supervised Action Segmentation.}
Fully supervised approaches rely on frame-level annotations during training. Earlier methods in this setup applied a sliding window approach with non-maximum suppression~\cite{rohrbach2012database,karaman2014fast}. Other approaches used context-free grammars~\cite{vo2014stochastic, pirsiavash2014parsing} 
or hidden Markov models (HMMs)~\cite{lea2016segmental, kuehne2016end, kuehne2020hybrid} to capture long-range dependencies. Recently, temporal convolutional networks (TCNs) with large receptive fields have been very successful in segmenting actions in long videos~\cite{lea2017temporal, lei2018temporal, farha2019ms, li2020ms}. Building on the success of these approaches, several approaches have been proposed to refine the TCN predictions using graph convolutional networks~\cite{huang2020improving}, boundary-aware pooling~\cite{wang2020boundary, ishikawa2020alleviating}, or hierarchical modeling~\cite{ahn2021refining}. In~\cite{gao2021global}, a neural network architecture search scheme is used to select the dilation factors for the TCN layers. Recently, \cite{yi2021asformer} proposed a transformer-based architecture for the temporal action segmentation task. In contrast to these approaches, our approach relies on a weaker level of supervision in the form of timestamps.

\paragraph{Weakly Supervised Action Segmentation.}
While fully supervised action segmentation approaches achieve very good results, they rely on dense frame-level annotations. As acquiring such annotations is time-consuming and expensive, many approaches have been proposed to train action segmentation models using weaker forms of supervision. A popular form of weak supervision are sequences of actions, which describe the order of actions in a video without any time information. The sequences of actions are also called transcripts. One of the first approaches~\cite{bojanowski2014weakly} for this setting used discriminative clustering. Other approaches addressed the segmentation task by aligning the network output to the transcript~\cite{ huang2016connectionist, chang2019d3tw, li2019weakly} or by generating pseudo ground truth using the Viterbi algorithm~\cite{kuehne2017weakly, ding2018weakly, richard2018neuralnetwork, kuehne2020hybrid}. At inference time, most of these approaches iterate over the transcripts seen during training and select the one with the highest alignment score to generate the final prediction. While such approaches generate good segmentation results, this comes at the cost of slow inference time. To alleviate this problem, \cite{mucon2021} combined a TCN-based backbone with a sequence-to-sequence model that predicts the transcript and segment lengths at inference time. To supervise the segment length output, an additional loss that enforces consistency between the sequence-to-sequence predictions and the TCN output is used. Recently, a gradient descent based approximation for the inference stage of the action segmentation has been proposed~\cite{fifa2021}. Another form of supervision that has recently emerged is based on video tags or action sets~\cite{richard2018action, fayyaz2020set, li2020set, li2021anchor}. In contrast to transcripts, set supervision provides only the set of actions that occur in the videos without any information regarding the order or how many times each action occurs. However, the performance of these approaches is still inferior compared to approaches that use transcripts for supervision. 

\paragraph{Timestamp Supervision for Action Segmentation.}
Timestamp supervision has been recently used for other activity understanding tasks. \cite{moltisanti2019action} proposed to sample frames around the annotated timestamps to train an action classifier based on a sampling function that is fitted to the classifier response. In the context of action localization, \cite{ma2020sf} proposed a mining strategy to sample action frames and background frames for training. In~\cite{lee2021learning}, this approach is extended in two ways. First, it uses a greedy approach to find for each action class the frames belonging to the action using sampled action frames and background frames as seeds. Second, it uses a contrastive loss assuming that there are at least two instances of the same action in a video. In the context of action segmentation, \cite{li2021temporal} has used timestamps for training and reported competitive results compared to fully supervised learning. For timestamp supervision, \cite{li2021temporal} proposed to annotate only a single frame for each action instance. To train an action segmentation model with this level of supervision, the model output and the timestamps are used to generate a labeling of an entire video, which is then used to update the model. While this approach has been effective, it assumes that all action instances are annotated by timestamps. By contrast, we relax this assumption in this paper and address the case where some actions have been missed.

\section{Robust Action Segmentation from
Timestamps}

\subsection{Timestamp Supervision for Action Segmentation}
Given an untrimmed video, action segmentation is the task of predicting the action label for each frame. The video is represented as a set of $D-$dimensional pre-extracted framewise features $X = \big( x_1, \cdots, x_T \big)$ with length $T$ and a network for action segmentation predicts the framewise probability estimates $\tilde{Y} = \big( \tilde{y}_1, \cdots, \tilde{y}_T \big)$, where $\tilde{y}_t \in [0, 1]^C$ represents the predicted probability for each action class and $C$ is the set of classes. We denote the ground truth labels for the video by $Y = \big( y_1, \cdots, y_T \big)$. 

While in case of full supervision the labels $Y$ are available for all frames during training, \cite{moltisanti2019action,li2021temporal} addressed the training setup where only one frame per action is annotated.          
The annotated frames are called timestamps, which we denote by \mbox{$P=(p_1, \cdots, p_N)$} and their corresponding labels by $y_{p_1}, \cdots, y_{p_N}$. We assume that the timestamps are already ordered, \ie, $$1 \le p_1 \le \cdots \le p_N \le T.$$ 
Timestamps are very sparse annotations of the training videos.   
For instance, less than 0.5\% of the frames are annotated for the 50Salads~\cite{stein2013combining} and Breakfast~\cite{kuehne2014language} datasets in case of timestamp supervision.

\begin{figure}[t]
   \centering
      \includegraphics[width=0.9\columnwidth]{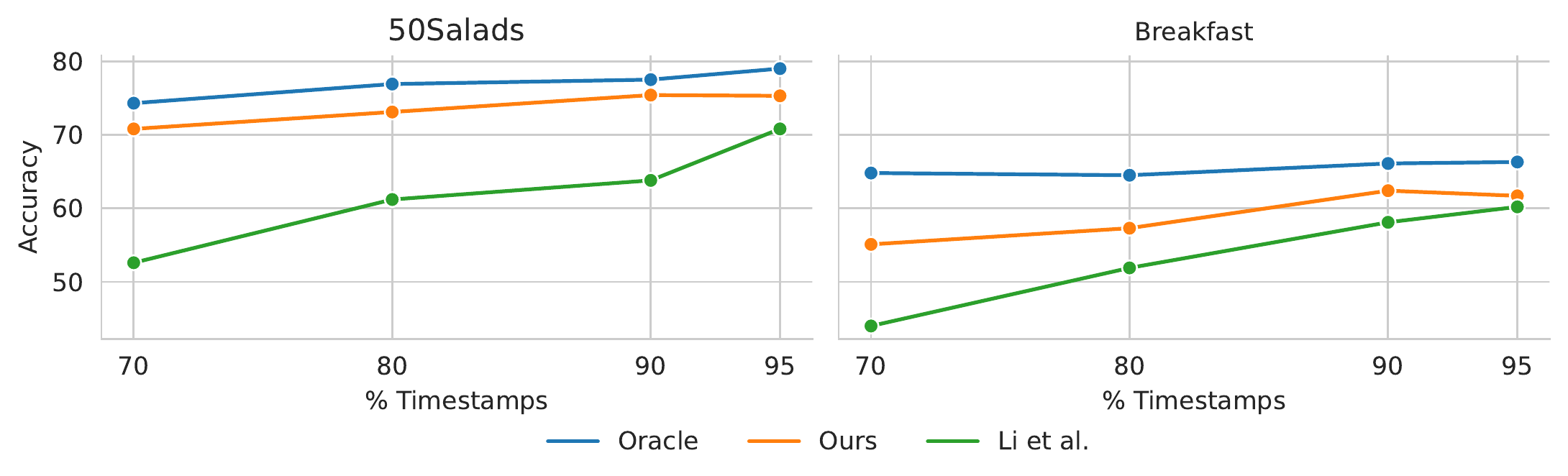}
   \caption{
       Accuracy of different action segmentation approaches trained using timestamp supervision with different percentages of annotated timestamps. Note that 100\% timestamps corresponds to one timestamp per action segment, which already covers less than 0.5\% of all frames. 
       The two plots show the framewise accuracy for the 50Salads~\cite{stein2013combining} and Breakfast dataset~\cite{kuehne2014language}, respectively. The `Oracle' is an upper bound of our approach and it corresponds to training in a fully supervised setting, but it ignores all frames of a ground-truth segment without timestamp. Our approach is more robust to missed timestamps than Li \etal~\cite{li2021temporal} and consistently achieves a higher accuracy. The gap between our approach and the approach of \cite{li2021temporal} increases as the percentage of missed timestamps increases. 
   }
   \label{fig:gap}
\end{figure}

While a naive approach for action segmentation just uses the timestamps for training the network, Li \etal~\cite{li2021temporal} proposed to generate a labeling of all frames from the timestamps. This is done by finding action boundaries between two consecutive timestamps in a forward and backward pass over the video, where each pass minimizes the variance of the framewise features that are separated by the action boundary. The approach, however, assumes that there is no other action between two consecutive timestamps. In other words, the approach assumes that the annotators did not miss any action during the labeling process. If an action is missed, the labeling generation process of \cite{li2021temporal} assigns the frames either to the previous or next timestamp, which is incorrect as it is illustrated in Figure~\ref{fig:teaser}. As a consequence, the approach is not very robust to missed annotations and the accuracy drastically decreases if only 80-90\% of the actions are annotated as shown in Figure~\ref{fig:gap}. While a recall of 90\% is realistic for larger datasets, 70\% is already very low given that 100\% annotated timestamps cover already less than 0.5\% of the frames.    

In this work, we, therefore, address the research question of how can we make the learning of networks for action segmentation more robust to missing timestamps.

\subsection{Robust Action Segmentation from Timestamps}

In order to make the labeling process, and thus the training of the network, more robust, we need to consider that some timestamps might have been missed during the annotation process. This can be achieved by labeling some frames as unknown as shown in Figure~\ref{fig:teaser}. Instead of detecting an action change between two consecutive timestamps as in \cite{li2021temporal}, we thus propose to expand the segment boundaries from the timestamps in each direction such that the network is confident that the frames within a segment belong to the same class as the timestamp and such that the number of frames that are unknown is minimized. The latter is important since there is otherwise no incentive to extend the segment boundaries.    

\begin{figure}[tb]
   \centering
      \includegraphics[width=0.55\columnwidth]{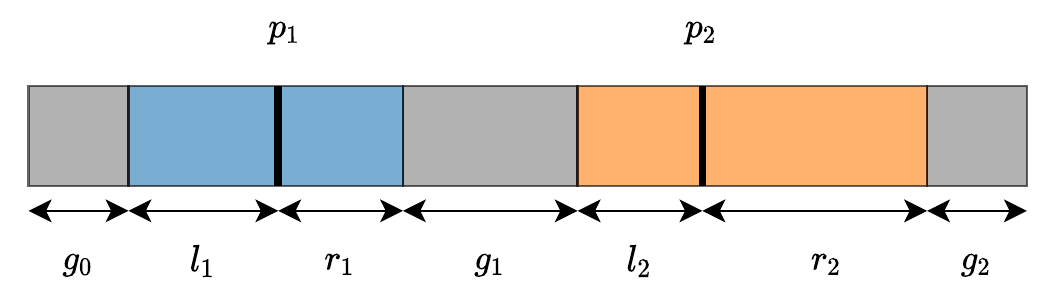}
   \caption{Given the annotated timestamps, we aim to identify the start and end of the actions that correspond to the timestamps. This is done by a joint optimization over all timestamps in a training video where we divide the frames between two consecutive timestamps $p_1$ and $p_2$ into three parts. The first part $r_1$ will take the same label as $p_1$ and the last part $l_2$ will take the same label as $p_2$. The gray colored frames $g_1$ will be labeled as unknown and ignored when training the network.
   }
   \label{fig:parameters}
\end{figure}

Figure~\ref{fig:parameters} shows the notation and illustrates the outcome of the labeling process for two timestamps. For each timestamp $p_i$, the corresponding action segment is defined by $[p_i - l_i, p_i + r_i]$. Between two timestamps $p_i$ and $p_{i+1}$, there can be $g_i$ frames that are labeled as unknown, \ie, $p_i + r_i + g_i + l_{i+1} = p_{i+1}$. We also allow the frames at the beginning and at the end of a video to be labeled as unknown, which we denote by $g_0$ and $g_N$, respectively. 

As \cite{li2021temporal}, we first train a network for action segmentation for a few epochs using only the annotated timestamps. During training, the network provides framewise probability estimates $\tilde{Y} = \big( \tilde{y}_1, \cdots, \tilde{y}_T \big)$ for all frames. We then optimize the parameters $l_i$, $r_i$, and $g_i$ such that on the one hand the predictions for the class label of the timestamp $y_{p_i}$ are highly confident within $[p_i - l_i, p_i + r_i]$ (colored regions in Figure~\ref{fig:parameters}) and such that on the other hand the number of unknown frames $g_i$ (gray regions in Figure~\ref{fig:parameters}) is as small as possible. This goal can be formulated for each video as the following constrained minimization problem:
\begin{equation}
   \begin{aligned}
   \{r^*_i, g^*_i, l^*_i\} =
   \underset{r_i, g_i, l_i}{\mathrm{argmin}}~&  \sum_{i=1}^N \bigg( \sum_{t = p_i - l_i}^{p_i + r_i} - \log \tilde{y}_t[y_{p_i}] \bigg) + \beta \sum_{i=0}^N g_i\\
   \textrm{s.t.} ~~ &  p_{i+1} - p_i = r_i + g_i + l_{i+1}\\
                    &  r_i \ge 0, ~~ g_i \ge 0, ~~ l_i \ge 0. \\
   \end{aligned}
   \label{eq:objective}
\end{equation}
The first constraint ensures that all frames are either labeled by the class of the associated timestamp or as unknown, and the last three constraints ensure that the number of frames for each part is not negative. Due to the constraints, the segments cannot overlap and the number of unknown frames is correctly counted. The hyper-parameter $\beta$ controls how strong the number of unknown frames should be penalized. We study the effect of this hyper-parameter in Section~\ref{sec:hparam}.

In order to optimize \eqref{eq:objective} efficiently, we can re-write the objective function as
\begin{equation}
    \begin{aligned}
    \sum_{i=1}^N \bigg(& \sum_{t = p_i - l_i}^{p_i + r_i} - \log \tilde{y}_t[y_{p_i}] \bigg) + \beta \sum_{i=0}^N g_i=\\
    \sum_{i=1}^N \bigg(& 
        \sum_{t = 1}^{T} - \log \tilde{y}_t[y_{p_i}] \mathcal{I} (t \vert p_i - l_i \le t \le p_i + r_i)
    \bigg)  \\
    + & \beta \sum_{t = 1}^{T} \bigg( 1 - \sum_{i=1}^N \mathcal{I} (t \vert p_i - l_i \le t \le p_i + r_i)
    \bigg) \\
    \end{aligned}
    \label{eq:objective_2}
\end{equation}
where $\mathcal{I} (t \vert p_i - l_i \le t \le p_i + r_i)$ is the indicator function with value $1$ if $t$ is within the left and right bounds of the timestamp and $0$ otherwise. 
As the $\mathcal{I}$ function is a non-differentiable function, we approximate it with a plateau function~\cite{moltisanti2019action,fifa2021}. Further details are given in the Appendix. 

To address the constraints of \eqref{eq:objective}, we re-parameterize $r_i$, $g_i$, and $l_{i+1}$ by $r'_i$, $g'_i$, and $l'_{i+1}$, respectively, such that
\begin{equation}
    z = \frac{(p_{i+1} - p_{i})\exp(z')}{\exp(r'_i)+\exp(g'_i)+\exp(l'_{i+1})}\quad\text{for}\quad z\in\{r_i,g_i,l_{i+1}\}. 
\end{equation}
In this way, we can optimize \eqref{eq:objective_2} for any value of $r'_i$, $g'_i$, and $l'_{i+1}$ without constraints since the re-parameterization ensures that $r_i \ge 0$, $g_i \ge 0$, $l_i \ge 0$, and $p_{i+1} - p_i = r_i + g_i + l_{i+1}$.
 
With the optimization problem becoming unconstrained and differentiable using the plateau function approximation, we can solve this optimization using gradient descent similar to FIFA~\cite{fifa2021}. 
This gradient-based optimization is solved only during training for each video independently. 
After the optimization, we use the generated labeling for each training video to continue the training of the network for action segmentation where the frames that are labeled as unknown are ignored.
More details are provided in the Appendix.

\label{sec:robust_gt_gen}

\section{Experiments}
\subsection{Datasets and Metrics}
\textbf{Datasets.} We evaluate our approach for action segmentation on the 50Salads~\cite{stein2013combining} and Breakfast~\cite{kuehne2014language} datasets. Furthermore, we provide additional results for the BEOID~\cite{damen2014you} and Georgia Tech Egocentric Activities (GTEA)~\cite{fathi2011learning} datasets for action localization. 

The 50Salads dataset consists of 50 videos of people preparing mixed salads. The average length of the videos is around 6.4 minutes. The dataset provides fine-granular annotations of 17 low-level activities such as \textit{peel cucumber} and \textit{mix ingredients}. For evaluation, we use five-fold cross-validation and report the average.

The Breakfast dataset contains 1712 videos of breakfast-related activities. The average length of the videos is around 2.3 minutes. The frames are annotated with 48 fine-grained action 
classes such as \textit{take bowl} and \textit{pour cereals}. For evaluation, we use the standard four
splits and report the average.

For both datasets, we use the timestamp annotations provided by~\cite{li2021temporal} and simulate missing annotations by randomly removing timestamps. Otherwise, the setup is the same as \cite{li2021temporal}. We use the same network for action segmentation and train it for 30 epochs using only the annotated timestamps. We then generate the labels for the training videos as described in Section \ref{sec:robust_gt_gen} and continue to train the network for 20 epochs on the labeled sequences. For frames that are labeled as unknown, the training loss is not computed, \ie, the frames are ignored for training the network. If not otherwise mentioned, we set the value of $\beta$ in \eqref{eq:objective_2} to $0.7$. 

More details regarding the optimization of \eqref{eq:objective_2} are provided in the Appendix.  

As in previous works for action segmentation, we report frame-wise accuracy (Acc), a score based on the segmental edit distance (Edit), and segmental F1 scores at overlapping thresholds $10\%,\ 25\%$ and $50\%$. For all metrics, a higher value is better.  

\subsection{Impact of Missing Annotations}
We first compare our approach with \cite{li2021temporal} for different percentages of annotated timestamps. While we evaluate our approach for a setting where we use all provided timestamps, \ie, 100\% of the timestamps, in Section~\ref{sec:allt}, we focus in this section on a setting where some actions have been missed during the annotation process. The results for the 50Salads and Breakfast datasets are reported in Table~\ref{tab:missingSegm}. Our approach is robust to missing annotations and outperforms the approach of \cite{li2021temporal} for all percentages and both datasets. The frame-wise accuracy of our approach drops by only $4.5\%$ on 50Salads and by $6.6\%$ on the Breakfast dataset when the percentage of annotated timestamps is reduced from $95\%$ to $70\%$. On the contrary, the accuracy of~\cite{li2021temporal} drops considerably by $18.2\%$ on 50Salads and by $16.2\%$ on the Breakfast dataset. The frame-wise accuracy is also plotted in Figure~\ref{fig:gap}, which shows that the gap between our approach and \cite{li2021temporal} increases as more timestamps are missing. Using $70\%$ of the timestamps for training, our approach outperforms \cite{li2021temporal} by $15.2\%$ and $11.5\%$ for the F1@25 metric on 50Salads and Breakfast, respectively. 

\begin{figure*}[t]
    \centering
    \includegraphics[trim={0 0 0 4pt},clip, width=0.9\textwidth]{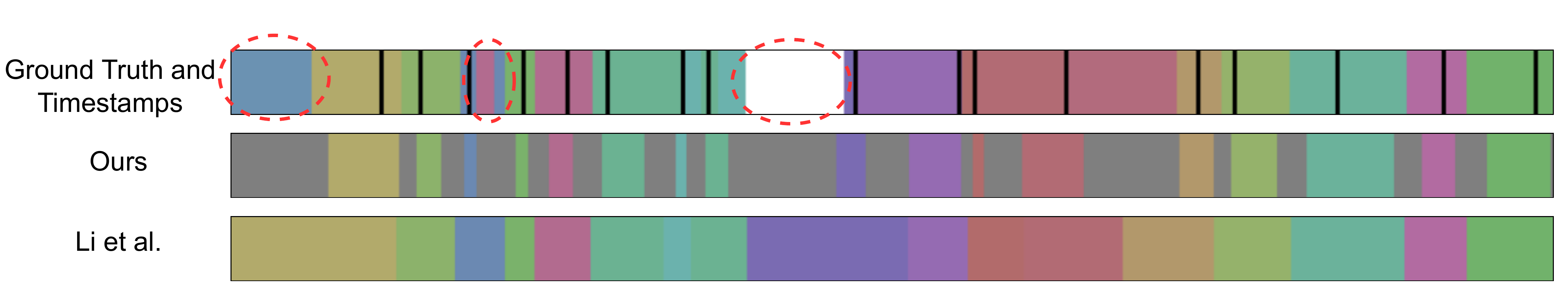}\\
    \caption{Qualitative results for comparing the generated labels by our approach and \cite{li2021temporal} on one video from the 50Salads dataset. The top row shows the ground truth frame-wise labels and the annotated timestamps as black lines. Note that only the 17 timestamps are given as annotation for training. 
    The missed actions are highlighted with red dashed ellipses. The labels that are generated by our approach from these 17 timestamps are shown in the second row, where the frames labeled as unknown are colored in gray. The labeling generated by~\cite{li2021temporal} is shown in the third row. While our approach correctly labels the frames of the missed actions correctly as unknown, \cite{li2021temporal} labels these frames wrongly. 
    }
    \label{fig:qualitative}
\end{figure*}

These results demonstrate that our proposed approach is much more robust to missing timestamp annotations than \cite{li2021temporal}. While the approach \cite{li2021temporal} generates dense labels, which results in wrong labels in the case of many missing timestamps, our approach explicitly handles missing annotations by ignoring some frames during training. This effect is also visible in the qualitative result shown in Figure~\ref{fig:qualitative}. While the approach of~\cite{li2021temporal} assigns wrong labels to the frames of actions that have been missed during the annotation process, our approach labels these frames as unknown.

Since the number of labeled actions decreases as more timestamps are missing, we also report the results for an oracle in Table~\ref{tab:missingSegm}. This oracle defines an upper bound that can be achieved by our method if $l_i$ and $r_i$ \eqref{eq:objective_2} are perfectly estimated for each annotated timestamp. To this end, we use the frame-wise ground-truth annotations of the training videos to get $l_i$ and $r_i$. Note that frames that belong to an action that has not been annotated by a timestamp are still ignored by the oracle. Therefore, the accuracy of the oracle also decreases as the percentage of annotated timestamps decreases. As shown in Figure~\ref{fig:gap}, the gap between our approach and its upper bound is quite constant for the 50Salads dataset while it increases slightly on Breakfast as the number of annotated timestamps decreases.

\begin{table}[tb]
\centering
    \resizebox{1.0\columnwidth}{!}{
    \begin{tabular}{cl | ccccc | ccccc}
            \hline
              \multirow{2}{*}{\% Timestamps} &  \multirow{2}{*}{Method}  &  \multicolumn{5}{c|}{Breakfast} & \multicolumn{5}{c}{50Salads}  \\
              \cline{3-12}
                &  & \multicolumn{3}{c}{F1@\{10, 25, 50\}} & Edit & Acc & \multicolumn{3}{c}{F1@\{10, 25, 50\}} & Edit & Acc  \\
             \hline
                  \multirow{3}{*}{95\%}  & Li \etal~\cite{li2021temporal} & 67.3  & 59.7  & 42.7  & 68.2  & 60.2 & 70.9  & 67.4  & 53.4  & 63.8  & 70.8   \\
                  & Ours & \textbf{70.2}  & \textbf{62.4}  & \textbf{44.8}  & \textbf{69.7}  & \textbf{61.7} & \textbf{72.9} & \textbf{69.6} & \textbf{57.5} & \textbf{64.2} & \textbf{75.3}   \\ 
                  & Oracle  & 71.0  & 65.3  & 51.4  & 70.2  & 66.3 & 74.8  & 72.0  & 64.1 & 67.9 & 79.0  \\
             \hline
                  \multirow{3}{*}{90\%} & Li \etal~\cite{li2021temporal} & 65.0  & 56.5  & 39.7  & 66.8  & 58.1 & 63.9  & 59.6  & 44.3  & 57.6  & 63.8 \\
                  & Ours & \textbf{69.8}  & \textbf{62.2}  & \textbf{44.5}  & \textbf{69.7}  & \textbf{62.4} & \textbf{70.0} & \textbf{65.1} & \textbf{55.2} & \textbf{62.1} & \textbf{75.4}  \\
                  & Oracle  & 69.0  & 63.0  & 49.1  & 68.6  & 66.1 & 73.9  & 71.6  & 62.5  & 66.9  & 77.5 \\
             \hline
                  \multirow{3}{*}{80\%} & Li \etal~\cite{li2021temporal} & 59.8  & 50.4  & 33.0  & 62.8  & 51.9  & 62.7  & 56.9  & 40.3  & 54.2  & 61.2 \\
                      & Ours & \textbf{67.3}  & \textbf{58.9}  & \textbf{40.8}  & \textbf{68.5}  & \textbf{57.3} & \textbf{70.9} & \textbf{67.8} & \textbf{53.7} & \textbf{61.4} & \textbf{73.1} \\
                      & Oracle  & 69.2  & 62.8  & 48.7  & 68.4  & 64.5 & 73.3  & 70.2 & 61.0 & 65.8 & 76.9 \\
              \hline
              \multirow{3}{*}{70\%} & Li \etal~\cite{li2021temporal} &  53.9  & 43.5 & 26.8  &  59.2  &  44.0 & 50.2  & 44.0  & 29.5  & 44.7  & 52.6  \\
                  & Ours &  \textbf{65.0}  & \textbf{55.0}  & \textbf{35.7}  &  \textbf{66.5}  &  \textbf{55.1} & \textbf{64.1} & \textbf{59.2} & \textbf{44.8} & \textbf{56.9} & \textbf{70.8} \\
                  & Oracle  &  69.5  &  63.2 & 49.0  &  69.0  &  64.8 & 67.4 & 63.3 & 53.2 & 59.0 & 74.3   \\

         \bottomrule
    \end{tabular}
    }
    \caption{Impact of missing timestamps on the Breakfast and 50Salads datasets.} \label{tab:missingSegm}
\end{table}
\subsection{Sensitivity to $\beta$}
\label{sec:hparam}
\begin{figure}[tb]
   \centering
      \includegraphics[width=0.85\columnwidth]{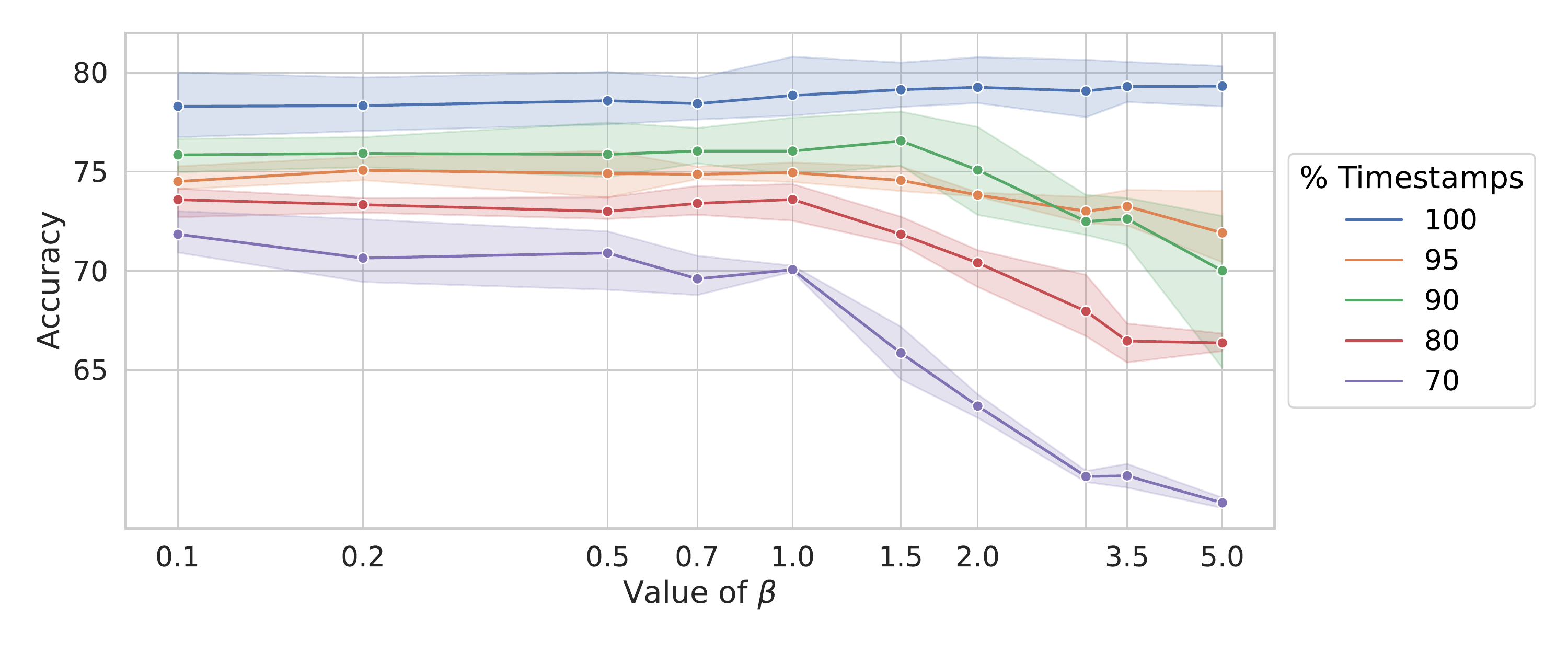}
   \caption{Impact of $\beta$ on the frame-wise accuracy for different percentages of annotated timestamps on the 50Salads dataset. For each experiment, the mean and standard deviation of three runs are shown.}
   \label{fig:empty_loss}
\end{figure}
In \eqref{eq:objective_2}, the number of frames that will be labeled as unknown is controlled by $\beta$. This is a hyper-parameter for our approach and has been set so far to $0.7$ in all experiments. 

In this section, we study the impact of $\beta$ on the accuracy, and Figure~\ref{fig:empty_loss} shows the frame-wise accuracy of our approach using different values for $\beta$. The results are reported for different percentages of annotated timestamps on the 50Salads dataset. We plot the mean and standard deviation of three runs for each setting.
As shown in the figure, lower values for $\beta$ tend to achieve better results if annotations are missing. The accuracy drops for large values of $\beta$, except for the setting where all timestamps are available (100\%). This is expected since with a lower percentage of annotations more frames should be labeled as unknown, which is achieved by lowering the value of $\beta$. When there are no missing timestamps, no frame should be labeled as unknown, and higher values of $\beta$ are better.
\subsection{Comparison without Missing Annotations}\label{sec:allt}
So far, we mainly considered the case where some actions have been missed since the goal of the proposed approach is to increase the robustness with respect to missing timestamp annotations. We finally evaluate our approach for the protocol used in~\cite{li2021temporal} where for each action a timestamp is provided. Table~\ref{tab:no_missing_annot} shows the results on both the 50Salads and the Breakfast dataset. While our approach is designed to handle missing annotations, it also works well when there are no missing timestamps. Our approach outperforms previous approaches on 50Salads and achieves competitive results on the Breakfast dataset. As discussed in Section~\ref{sec:hparam}, if we know that there are no missing timestamps, then we can directly use a higher value of $\beta$ (for instance $\beta = 2$). By increasing $\beta$, we encourage our approach to decrease the number of frames that are labeled as unknown, which improves the accuracy. For completeness, we also compare to fully supervised approaches and approaches that are trained with less supervision.    

\begin{minipage}[t]{0.95\columnwidth}
\vspace{3mm}
\setlength\tabcolsep{2pt}
\begin{minipage}[t]{0.6\columnwidth}
\resizebox{\columnwidth}{!}{%
      \begin{tabular}{cl  ccccc | ccccc}
                \hline
               \multirow{2}{*}{Supervision} &  \multirow{2}{*}{Method}  &  \multicolumn{5}{c|}{Breakfast} & \multicolumn{5}{c}{50Salads}  \\
              \cline{3-12}
                &  & \multicolumn{3}{c}{F1@\{10, 25, 50\}} & Edit & Acc & \multicolumn{3}{c}{F1@\{10, 25, 50\}} & Edit & Acc  \\
             \hline
         \multirow{7}{*}{Full} & 
         MS-TCN~\cite{farha2019ms} & 52.6 & 48.1 & 37.9 & 61.7 & 66.3 & 76.3 & 74.0 & 64.5 & 67.9 & 80.7  \\
         & MS-TCN++~\cite{li2020ms} & 64.1 & 58.6 & 45.9 & 65.6 & 67.6 & 80.7 & 78.5 & 70.1 & 74.3 & 83.7  \\
         & BCN~\cite{wang2020boundary} & 68.7 & 65.5 & 55.0 & 66.2 & 70.4 & 82.3 & 81.3 & 74.0 & 74.3 & 84.4  \\
         & ASRF~\cite{ishikawa2020alleviating} & 74.3 & 68.9 & 56.1 & 72.4 & 67.6 & 84.9 & 83.5 & 77.3 & 79.3 & 84.5  \\
         & FIFA~\cite{fifa2021}  & 75.5  & 70.2  & 54.8  & 78.5  &  68.6 & - & - & - & - & -  \\
         & MuCon~\cite{mucon2021}  & 73.2  & 66.1  & 48.4  & 76.3  & 62.8 & - & - & - & - & -   \\
         & ASFormer~\cite{yi2021asformer}  & 76.0  & 70.6  & 57.4  & 75.0  &73.5 & 85.1  & 83.4  & 76.0  & 79.6  &85.6   \\
         \midrule
          \multirow{4}{*}{Timestamps} & 
         Plateau~\cite{moltisanti2019action}$^*$ & 65.5 & 59.1 & 43.2 & 65.9 & \underline{63.5} & 71.2 & 68.2 & 56.1 & 62.6 & 73.9 \\
         & Li \etal~\cite{li2021temporal} &  \underline{70.5} & \underline{63.6} & \textbf{47.4} & \underline{69.9} & \textbf{64.1} & 73.9  & 70.9  & 60.1  &  66.8 & 75.6  \\
         & Ours & 67.0   & 60.0  & 43.8  & 66.7  & 60.8 & \underline{75.2} & \underline{72.1} & \underline{61.5} & \underline{67.7} & \underline{78.4}  \\
         & Ours ($\beta = 2$) & \textbf{71.5}   & \textbf{64.3}  & \underline{47.3}  & \textbf{70.9}  & 62.9  &\textbf{77.0} & \textbf{74.2} & \textbf{62.2} & \textbf{69.8} & \textbf{79.3} \\
         \bottomrule
      \end{tabular}
    }
\end{minipage}
\begin{minipage}[t]{0.43\columnwidth}
    \resizebox{0.9\columnwidth}{!}{%
      \begin{tabular}{cl  cc}
         \hline
         \multirow{2}{*}{Supervision} & \multirow{2}{*}{Method} & Breakfast & 50Salads \\
         \cline{3-4}
          & & \multicolumn{2}{c}{Acc} \\
         \hline
         \multirow{7}{*}{Transcripts} 
         & FIFA~\cite{fifa2021} & 51.3 & - \\
         & CDFL~\cite{li2019weakly} & 50.2 & 54.7\\
         & HMM-RNN~\cite{richard2017weakly} & - & 45.5 \\
         & MuCon~\cite{mucon2021} & 48.5 & - \\
         & D$^{3}$TW~\cite{chang2019d3tw} & 45.7 & -  \\
         & NN-Viterbi~\cite{richard2018neuralnetwork} & 43.0 & 49.4  \\
         & TCFPN~\cite{ding2018weakly}  & 38.4 & - \\
         \hline
         \multirow{3}{*}{Sets} & 
         SCT~\cite{fayyaz2020set}  & 30.4 & - \\
         & SCV~\cite{li2020set}  & 30.2 & - \\
         & Action Sets~\cite{richard2018action} & 23.3 & - \\
         \bottomrule
      \end{tabular}
    }
   \end{minipage}
\captionof{table}{Results for timestamp annotations (100\%) on the Breakfast and 50Salads datasets. For completeness, we also compare to fully supervised approaches and approaches that are trained with less supervision (transcripts or action sets). $^*$ are results reported in  \cite{li2021temporal}.}
\label{tab:no_missing_annot}
\end{minipage}

\subsection{Action Localization}

While our approach is designed for action segmentation, it can also be applied to the action localization task. We report the results on the BEOID~\cite{damen2014you} and the Georgia Tech Egocentric Activities (GTEA)~\cite{fathi2011learning} dataset. 
Following~\cite{li2021temporal}, we use our label generation approach to train an action localization model using the human-annotated timestamps on the GTEA and BEOID datasets provided by~\cite{ma2020sf}. Results for both datasets are shown in Table~\ref{tab:sfnetmap}. Our approach outperforms~\cite{li2021temporal} and \cite{ma2020sf} by a large margin and achieves competitive results compared to~\cite{lee2021learning}, which is a specific approach for action localization. 

\begin{table}[h]
\setlength{\tabcolsep}{4pt}
   \centering
      \begin{tabular}{@{\hskip .2in}l@{\hskip .4in}c cccc}
         \toprule
         mAP@IoU & 0.1 & 0.3 & 0.5 & 0.7 & Avg  \\
         \midrule
         \emph{\textbf{GTEA}} \\
         \midrule
         SF-Net~\cite{ma2020sf} & 58.0 & 37.9 & 19.3 & 11.9& 31.0 \\
         Li \etal~\cite{li2021temporal}  & 60.2 & 44.7 & 28.8 & 12.2 & 36.4 \\
         Lee~\etal~\cite{lee2021learning}  & \textbf{63.9} &  \textbf{55.7} & \underline{33.9} & \textbf{20.8} & \textbf{43.5} \\
         Ours  & \underline{63.6} & \underline{54.1} & \textbf{36.4} & \underline{20.3} & \underline{43.4} \\
         \midrule
         \emph{\textbf{BEOID}} \\
         \midrule
         SF-Net~\cite{ma2020sf} & 62.9 & 40.6 & 16.7 & 3.5 & 30.1 \\
         Li \etal~\cite{li2021temporal} & 71.5 & 40.3 & 20.3 & 5.5 & 34.4 \\
         Lee~\etal~\cite{lee2021learning}  & \underline{76.9} & \underline{61.4} & \textbf{42.7} & \textbf{25.1} & \underline{51.8} \\
         Ours  & \textbf{79.0} & \textbf{68.3} & \underline{42.1} & \underline{17.2} & \textbf{52.9} \\
         \bottomrule
      \end{tabular}
   \caption{Comparison with other approaches for action localization with timestamp supervision on the GTEA and BEOID dataset.}
   \label{tab:sfnetmap}
\end{table}


\section{Conclusion}
In this paper, we proposed a robust action segmentation approach for timestamp supervision. In contrast to the previous method~\cite{li2021temporal} that assumes that all actions are annotated by a timestamp, we proposed a label generation method that handles missing timestamp annotations and is thus more suitable for real-world applications. While our approach achieves competitive results when all actions are annotated, it is much more robust to missing annotations. Our approach achieves considerably higher accuracy, edit score, and F1 score compared to~\cite{li2021temporal} when actions have been missed during the annotation process.

\paragraph{Acknowledgement}
The work has been supported by the Deutsche Forschungsgemeinschaft (DFG, German Research Foundation) GA 1927/4-2 (FOR 2535 Anticipating Human Behavior), MKW NRW iBehave, and the ERC Consolidator Grant FORHUE (101044724). 

\newpage

\appendix
\section*{Appendix}
We provide further details of the optimization, additional ablation studies, and report the runtime.  

\section{Optimization}\label{sec:opt}
\begin{figure}[h]
    \centering
       \includegraphics[width=0.60\columnwidth]{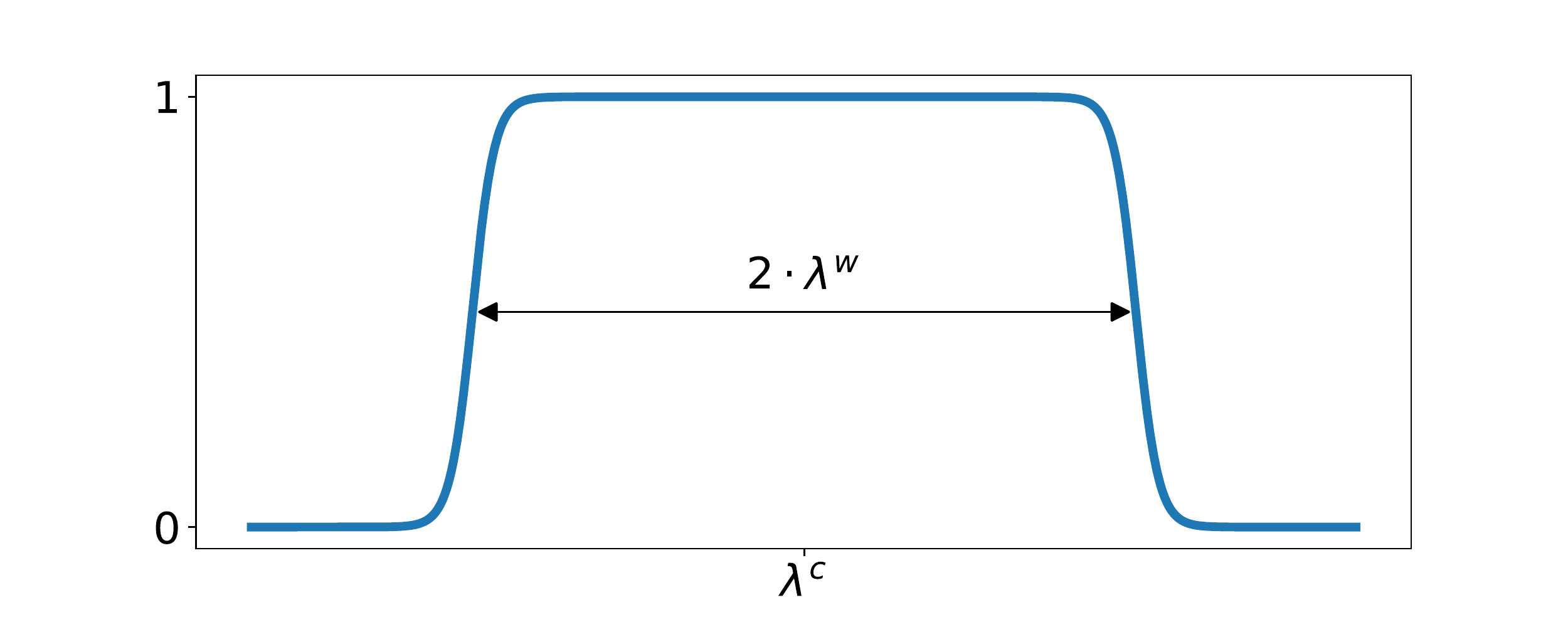}
    \caption{The plateau function (\ref{eq:plateau}) with center parameter $\lambda^c$ and width parameter $\lambda^w$.}
    \label{fig:plateau_function}
\end{figure}

As discussed in the paper, we optimize the objective: 
\begin{equation}
    \begin{aligned}
    \sum_{i=1}^N \bigg(& 
        \sum_{t = 1}^{T} - log \tilde{y}_t[y_{p_i}] \mathcal{I} (t \vert p_i - l_i \le t \le p_i + r_i)
    \bigg)  \\
    + & \beta \sum_{t = 1}^{T} \bigg( 1 - \sum_{i=1}^N \mathcal{I} (t \vert p_i - l_i \le t \le p_i + r_i)
    \bigg) \\
    \end{aligned}
    \label{eq:objective_2}
\end{equation}
As the indicator function $\mathcal{I}$ is a non-differentiable function, we replace it with the differentiable plateau function from \cite{moltisanti2019action,fifa2021}.
The plateau function shown in Figure~\ref{fig:plateau_function} is defined by
\begin{equation}
    \begin{aligned}
        f(&t \vert \lambda^c, \lambda^w, \lambda^s) =  \frac{1}{(e^{\lambda^s(t-\lambda^c-\lambda^w)} + 1)(e^{\lambda^s(-t+\lambda^c-\lambda^w)} + 1)}.
    \end{aligned}
    \label{eq:plateau}
\end{equation}
It defines a window of size $2 \lambda^w$ at the center $\lambda^c$. The parameter $\lambda^s$ of the plateau function controls the sharpness of the transition from $0$ to $1$.

For optimization, we replace the indicator function $\mathcal{I}$ by the plateau function $f$:
\begin{align}
    \mathcal{I} (t \vert p_i - l_i \le t \le p_i + r_i) =  f(t \vert \lambda^{c_i}, \lambda^{w_i}, \lambda^s)
\end{align}
where $\lambda^{c_i} = p_i + \frac{r_i - l_i}{2}$, $\lambda^{w_i} = \frac{r_i + l_i}{2}$, and $\lambda^s=0.025$ is fixed.
Equation (\ref{eq:objective_2}) is thus re-written as
\begin{equation}
    \begin{aligned}
    \sum_{i=1}^N \bigg( 
        \sum_{t = 1}^{T} - log \tilde{y}_t[y_{p_i}] f(t \vert \lambda^{c_i}, \lambda^{w_i}, \lambda^s)
    \bigg) + \beta \sum_{t = 1}^{T} \bigg( 1 - \sum_{i=1}^N f(t \vert \lambda^{c_i}, \lambda^{w_i}, \lambda^s) \bigg).
    \end{aligned}
    \label{eq:objective_final}
\end{equation}
For the gradient descent based optimization of \eqref{eq:objective_final}, we initialize $r_i, g_i, l_{i+1}$ uniformly, \ie, $r_i=g_i=l_{i+1}$ and $r_i+g_i+l_{i+1}=p_{i+1}-p_i$. We optimize \eqref{eq:objective_final} for 30 iterations using the Adam optimizer with a learning rate of 0.03.


\begin{table}[t]
\setlength\tabcolsep{4pt} 
  \centering
      \begin{tabular}{clccccc}
         \toprule
         \% Segments & Method  &  \multicolumn{3}{c}{F1@\{10, 25, 50\}} & Edit & Acc  \\
         \midrule
         95\%      & Uniform-2 & 63.1  & 56.4  & 37.8  & 58.8  &59.5    \\
                  & Uniform-3 & 63.4  & 58.5  & 40.8  & 56.9  & 63.5   \\
                  &Timestamps only  & 59.9  & 55.2  & 45.6  & 49.6  & 71.5\\
                  & Ours & \textbf{72.9} & \textbf{69.6} & \textbf{57.5} & \textbf{64.2} & \textbf{75.3} \\
         \midrule
         90\%     & Uniform-2 & 60.8  & 53.0  & 34.7  & 56.0  & 56.1   \\
                  & Uniform-3 & 62.0  & 56.3  & 39.2  & 56.1  & 61.5   \\
                  &Timestamps only  & 55.4  & 51.4  & 40.2  & 46.0  & 69.6\\
                  & Ours & \textbf{70.0} & \textbf{65.1} & \textbf{55.2} & \textbf{62.1} & \textbf{75.4}  \\
        \midrule
         80\%     & Uniform-2 & 56.2  & 49.3  & 32.1  & 51.1  & 56.3   \\
                  & Uniform-3 & 59.6  & 52.5  & 35.3  & 54.4  & 59.7   \\
                  &Timestamps only  & 55.1  & 50.8  & 39.6  & 44.8  & 66.2 \\
                  & Ours & \textbf{70.9} & \textbf{67.8} & \textbf{53.7} & \textbf{61.4} & \textbf{73.1}  \\ 
        \midrule
         70\%    & Uniform-2 & 42.2  & 36.0  & 19.0  & 40.1  & 45.8   \\
                  & Uniform-3 & 48.8  & 43.2  & 28.5  & 46.0  & 54.1   \\
                  &Timestamps only  & 46.6  & 41.4  & 30.2  & 39.3  & 60.0\\
                  & Ours & \textbf{64.1} & \textbf{59.2} & \textbf{44.8} & \textbf{56.9} & \textbf{70.8} \\
 
         \bottomrule
      \end{tabular}
  \caption{Comparison with different baselines on the 50Salads dataset.}
  \label{tab:50Salads_baselines}
\end{table}

\section{Additional Ablation Studies}
\subsection{Comparison with Baselines}
We compare our optimization approach with a few baselines. The first baseline uses only the annotated timestamps for training and ignores all the frames in between, which is denoted by ``Timestamps only". The second baseline ``Uniform-2" divides the frames between the timestamps equally into two segments and assigns labels to each frame based on the label of the nearest timestamp. Whereas in the last baseline ``Uniform-3", the frames between timestamps are divided into three equally sized segments. In this baseline, only the first and last segments are labeled by the corresponding timestamp and the middle segment is ignored during training.
Results for our approach and the baselines on the 50Salads dataset are shown in Table~\ref{tab:50Salads_baselines}. Our approach outperforms all baselines.


\subsection{Impact of Initialization}
As discussed in Section \ref{sec:opt}, we initialize $r_i, g_i, l_{i+1}$ uniformly (Uniform-3). To analyse the impact of the initialization of the optimization, we compare it to another initialization where we set $l_{i}$ and $r_i$ to 3 seconds and $g_i=p_{i+1}-p_i-r_i-l_{i+1}$. Table~\ref{tab:50Salads_init} shows the results of the uniform initialization compared to the initialization based on a fixed duration. The uniform initialization scheme performs better.
\begin{table}[t]
\setlength\tabcolsep{4pt} 
   \centering
      \begin{tabular}{clccccc}
         \toprule
         \% Segments & Initialization  &  \multicolumn{3}{c}{F1@\{10, 25, 50\}} & Edit & Acc  \\
         \midrule
         95\%      & Fixed (3 sec) &  69.7  & 66.9   & 55.3   & 62.4    & 73.3    \\
                  & Uniform &  \textbf{72.9} & \textbf{69.6} & \textbf{57.5} & \textbf{64.2} & \textbf{75.3}     \\
         \midrule
         90\%     & Fixed (3 sec) & 68.4    & \textbf{65.7}   & \textbf{55.3}   &58.5    & 72.9    \\
                  & Uniform &  \textbf{70.0} & 65.1 & 55.2 & \textbf{62.1} & \textbf{75.4}  \\
        \midrule
         80\%     & Fixed (3 sec) & 66.2   & 63.1   & 50.7   & 57.6   &    71.1 \\
                  & Uniform &  \textbf{70.9} & \textbf{67.8} & \textbf{53.7} & \textbf{61.4} & \textbf{73.1}     \\
        \midrule
         70\%    & Fixed (3 sec) & 62.0   & 58.5    &  44.3  & 53.5   &  67.0   \\
                 & Uniform &  \textbf{64.1} & \textbf{59.2} & \textbf{44.8} & \textbf{56.9} & \textbf{70.8}    \\
 
         \bottomrule
      \end{tabular}
   \caption{Impact of initialization on the 50Salads dataset.}
   \label{tab:50Salads_init}
\end{table}

\begin{table}[t]
\setlength\tabcolsep{4pt} 
   \centering
      \begin{tabular}{llccccc}
         \toprule
         Method & Timestamps  &  \multicolumn{3}{c}{F1@\{10, 25, 50\}} & Edit & Acc  \\
         \midrule
         Li \etal~\cite{li2021temporal}     & Start frame & 49.7  & 36.8  & 14.8  & 49.8  & 41.5    \\
                  & Center frame & 69.5  & 65.6  & 48.5  & 61.8  & 66.6    \\
                  & Gaussian & 67.1  & 62.5  & 45.4  & 58.0  & 66.3    \\
                  & Uniform & 63.9  & 59.6  & 44.3  & 57.6  & 63.8  \\ 
         \midrule
         Ours     & Start frame & 54.1  & 40.7  & 17.3  & 52.8  &  44.8   \\
                  & Center frame & 71.5  & 68.9  & 56.9  & 63.2  &  72.4   \\
                  & Gaussian  & 70.8  & 67.2  & 55.4  & 62.3 & 71.9   \\
                  & Uniform & 70.0 & 65.1 & 55.2 & 62.1 & 75.4  \\
         \bottomrule
      \end{tabular}
   \caption{Results for different setups for providing timestamps. We use 90\% of the timestamps on the 50Salads dataset.}
   \label{tab:50Salads_timestamps_selection}
\end{table}

\subsection{Evaluation of Different Timestamps Selection Strategies}
The timestamps provided by \cite{li2021temporal} follow a uniform distribution. We also analyze the performance if the timestamps follow a Gaussian distribution. To this end, we randomly sampled a timestamp for each ground-truth action from a Gaussian distribution using the center of the action as the mean and half of the duration of the action as the standard deviation. If the sample is outside the action, we set it to the start or end frame of the action, respectively. We also consider the case where the timestamps are at the center of each action and the worst case where all timestamps are at the beginning of each action. As pointed out in the supplemental material of  \cite{li2021temporal}, humans would not annotate the start frame since it is more ambiguous. Table~\ref{tab:50Salads_timestamps_selection} shows that our approach outperforms \cite{li2021temporal} regardless of how the annotated timestamps are provided.

Finally, we evaluate a setup where action segments that are difficult to recognize by the network are more likely to be missed. To identify these segments, we trained a model using all timestamps for 30 epochs and used it to compute the average probability of the correct class for each ground-truth action segment. We set the sampling probability of a timestamp proportional to the inverse
of the class probability of the corresponding ground-truth segment, i.e., timestamps with a low prediction probability are less likely to be sampled. We then sampled 95\% of the action segments without replacement. We report the results in Table \ref{tab:sampling_startegy}.

\begin{table}[t]
\centering
    \begin{tabular}{l | ccccc}
            \hline
              \multirow{2}{*}{Method}  & \multicolumn{5}{c}{50Salads}  \\
              \cline{2-6}
                & \multicolumn{3}{c}{F1@\{10, 25, 50\}} & Edit & Acc  \\
             \hline
                  Li \etal~[28]  & 64.7  & 60.1  & 47.1  & 57.1  & 67.5   \\
                                                  Ours &  \textbf{65.3} & \textbf{61.1} & \textbf{49.8} & \textbf{58.3} & \textbf{71.0}   \\ 
                                                  Oracle  &  74.2  & 72.4  & 62.6 & 64.8 & 75.8  \\

         \bottomrule
    \end{tabular}
    \caption{Results if segments that are difficult to recognize by the network are missed. The results are reported on split 1 of the 50Salads dataset for 95\% of the timestamps.} \label{tab:sampling_startegy}
\end{table}

\subsection{Unknown Frames}
\begin{figure}[t!]
    \centering
    \includegraphics[width=.85\columnwidth]{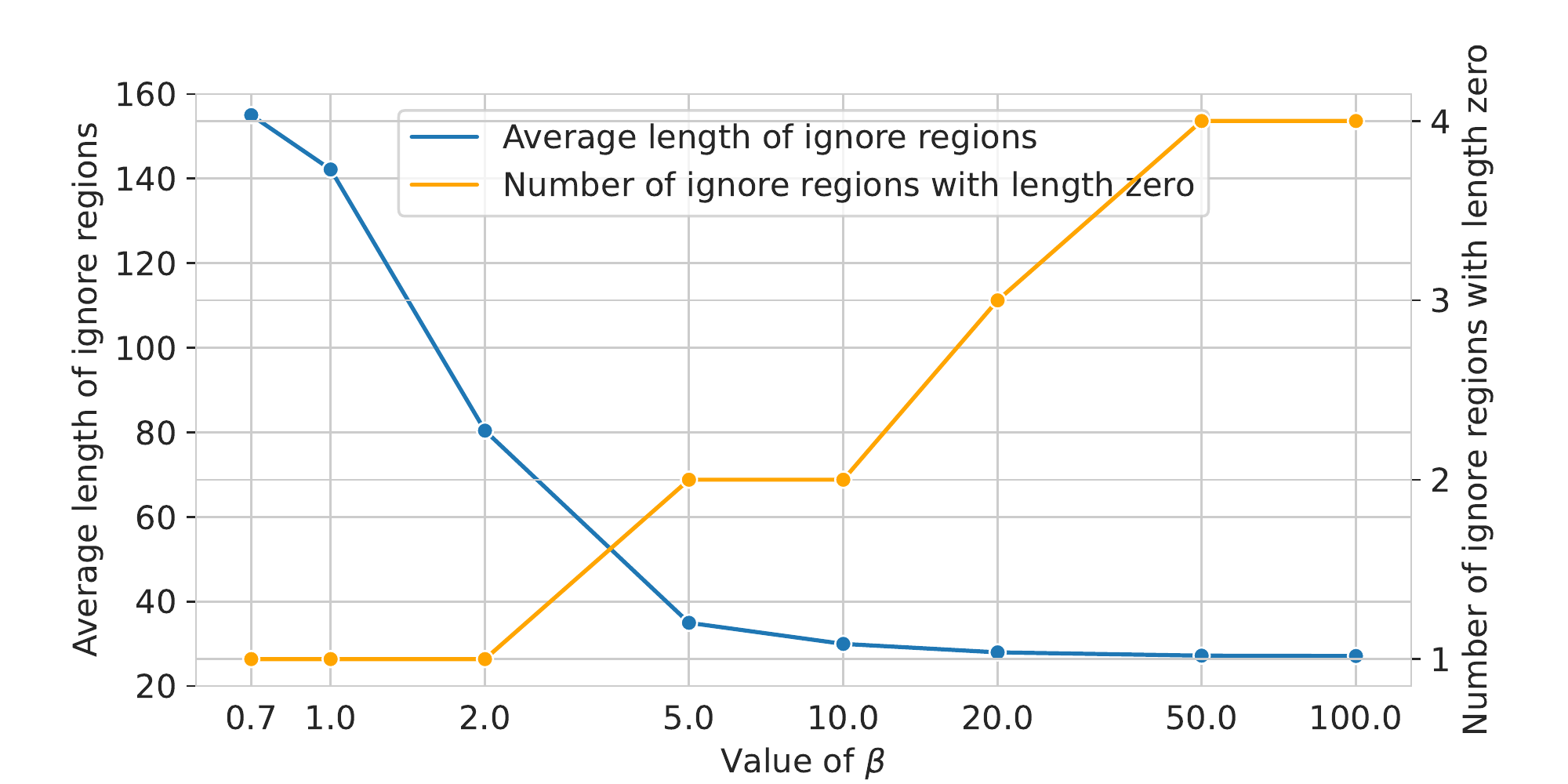}
    \caption{Average length of the ignore regions (average value of $g_i$) and number of the ignore regions with length 0 ($g_i=0$) for different values of $\beta$. The numbers are reported for the training set of split 1 of the 50Salads dataset.}
    \label{fig:empty_ignore}
\end{figure}

In the paper, we have already analyzed the impact of $\beta$ on the accuracy. Figure~\ref{fig:empty_ignore} shows the average value of $g_i$ (average length of an ignore region) and how often $g_i=0$ (length zero) for different values of $\beta$. The results are reported for the training set of split 1 of the 50Salads dataset.
As expected, the average size of $g_i$ decreases as the value of $\beta$ increases. Furthermore, we see that, even for large values of $\beta$, it occurs rarely that $g_i=0$. This is desirable since there is usually a transition between two actions that should not be labeled by any of the two actions.      

\section{Runtime Comparison}
Our proposed approach for generating labels from timestamps is not only more robust than \cite{li2021temporal}, but it is also much faster. We measured the wall clock time for  the whole training set of split 1 of the 50Salads dataset. While \cite{li2021temporal} requires 116 seconds to generate the labels, our approach requires only 1.7 seconds, which is 68 times faster.

{\small
\bibliography{references}
}

\end{document}